\documentclass[letterpaper]{article} % DO NOT CHANGE THIS
\usepackage[preprint]{aaai2027}  % DO NOT CHANGE THIS
\usepackage[hyphens]{url}  % DO NOT CHANGE THIS
\usepackage{graphicx} % DO NOT CHANGE THIS
\usepackage{natbib}  % DO NOT CHANGE THIS AND DO NOT ADD ANY OPTIONS TO IT
\usepackage{caption} % DO NOT CHANGE THIS AND DO NOT ADD ANY OPTIONS TO IT
\usepackage{algorithm}
\usepackage{algorithmic}
\usepackage{multirow}
\usepackage{rotating}
\usepackage{booktabs}
\usepackage{url}
\usepackage{tabularx}
\usepackage{array}
\usepackage{todonotes}
\usepackage{amsmath}

\usepackage{newfloat}
\usepackage{listings}
\DeclareCaptionStyle{ruled}{labelfont=normalfont,labelsep=colon,strut=off} % DO NOT CHANGE THIS
\floatstyle{ruled}
\newfloat{listing}{tb}{lst}{}
\floatname{listing}{Listing}

\usepackage{booktabs}

\usepackage{pifont}

\newcommand{\cmark}{\ding{51}}
\newcommand{\xmark}{\ding{55}}
\newcommand{\pmark}{$\circ$}

\title{Towards Stream Learning on Embedded Systems: Benchmarking the Memory Consumption of Stream Learning Methods}
\author{
    Sebastian Buschjäger\textsuperscript{\rm 1}\corresponding,
    Nuwan Gunasekara\textsuperscript{\rm 2},
    Heitor Murilo Gomes\textsuperscript{\rm 3}
}
\affiliations{
    \textsuperscript{\rm 1}Lamarr Institute, Dortmund, Germany, sebastian.buschjaeger@tu-dortmund.de \\
    \textsuperscript{\rm 2}Lamarr Institute,  Halmstad University, Sweden, nuwan.gunasekara@hh.se \\
    \textsuperscript{\rm 3}Victoria University of Wellington, New Zealand, heitor.gomes@vuw.ac.nz\\
    
}

\begin{document}

\maketitle

\begin{abstract}
Stream learning is commonly evaluated through predictive performance and adaptation to concept drift. However,  sustained operation of a stream learner also requires predictable and bounded resource usage even on long streams. This requirement becomes even more critical when learning moves from servers to near-sensor embedded systems where memory and processing are scarce resources. In state-of-the-art stream learning, however, we perceive a strong focus on concept drift adaptation, whereas resource usage is often an evaluation byproduct. To close this gap, we benchmark seven representative stream classifiers on 13 real and synthetic streams under model-size budgets from 128\,KiB to approximately 8\,MiB. Our benchmark comprises a total of 6,463 experiments. We measure failure-aware accuracy, peak model size, time to budget exhaustion, and prediction-plus-update latency. The results reveal two distinct resource failure modes. Adaptive ensembles can exceed small budgets almost immediately because of their initial footprint, even when their size remains stable thereafter. Incremental trees can fit initially but grow throughout a long stream, with HoeffdingTrees (HT) and Extremely Fast Decision Trees (EFDT) increasing by median factors of 7.37 and 5.87. This growth is often, but not invariably, accompanied by increasing latency.
Explicitly compact methods remain the only viable option under the smallest budgets, but are usually overtaken as larger budgets make adaptive ensembles competitive. These findings show that being incremental or drift-aware does not make a learner automatically resource-bound. Hence, many state-of-the-art methods are only partially applicable in embedded systems or for long-running systems. We therefore call on the stream-learning community to make bounded resource usage a first-class design objective alongside drift adaptation, and propose concrete steps toward this goal, including an API through which stream learners can explicitly expose and respect resource budgets.
% briefly discuss potential solutions 
% We conclude the paper with a call to action in which we 
% We conclude that embedded stream learning requires not only budget-aware algorithms and long-horizon evaluation, but also software interfaces that expose memory measurement and enforcement as first-class operations.
\end{abstract}

%     \link{Code}{See~supplementary~material}
%     \link{Results}{}
% %     \link{Extended version}{https://aaai.org/example/extended-version}

\section{Introduction}
Data stream learning addresses settings in which examples arrive continuously, predictions must be made before labels are observed, and models are updated incrementally under limited computational resources. This setting is especially natural for near-sensor processors such as the microcontroller units (MCUs) used in wearables, industrial sensors, environmental monitors, and low-power medical devices. See Table~\ref{tab:embedded-memory-regimes} for an overview of devices. On such devices, memory is typically measured in kilobytes to a few megabytes rather than gigabytes. The hardware therefore enforces the classical streaming assumptions: each example is processed once, historical data cannot be retained, and the learner must remain memory bounded. This is in stark contrast to larger embedded Linux boards and edge servers, which typically have gigabytes of memory and thus allow for buffering, replay, or offloading.  %In this paper, we focus on the MCU-class regime, where a stream learner that grows in size over time can become undeployable even if every individual update is incremental.
\begin{table}
\centering 
\caption{Representative embedded platforms. RAM denotes total platform memory, of which only a fraction is available to the learner after accounting for the runtime, application, buffers, and operating system. Values are taken from reference documentation of each device.} 
\label{tab:embedded-memory-regimes} 
\footnotesize \begin{tabular}{lr} 
\toprule Platform Examples & Typical RAM \\
\midrule 
Arduino Uno, Mega & 1--8 KB \\
Arduino Uno R4, Raspberry Pi Pico 2 & 32--512 KB \\
ESP32-S3, nRF54H20 & 0.5--4 MB\\ %, %STM32N6~\cite{esp32s3,nrf54h20,stm32n6} \\
i.MX RT1170, STM32H7/N6& 1--4 MB  \\
Raspberry Pi Zero 2 W, Raspberry Pi 5 & 0.5--16 GB \\
NXP i.MX 93, TI AM62A + NPU & 2--4 GB  \\
Jetson Orin Nano / AGX Orin & 8--64 GB \\
\bottomrule 
\end{tabular} 
\end{table}

A stream learner is expected not only to adapt online but also to remain usable over long time horizons. This requires bounded model size, stable update and prediction time, and robust predictive performance even after millions of updates. These requirements were already explicit in early stream-mining systems. For example, Bifet et al.\ describe MOA as enabling experiments on tens of millions of examples under explicit memory limits~\cite{bifet2018machine}. %Processing data incrementally is therefore not sufficient if the learner continues to grow with the number of observed examples. A stream learner should remain deployable after long sequences of updates, not only accurate during the early part of a stream.
However, much of the empirical practice in stream learning focuses on adaptation to concept drift and predictive accuracy as opposed to sustained resource behavior. Hoeffding-style trees \cite{domingos2000mining, bifet2009adaptive, manapragada2018extremely}, and their ensembles~\cite{ARF_gomes2017adaptive, SRP_Gomes2019streaming} can refine their structure as data arrive and react to change. The same mechanisms can also cause continued growth under noise, recurring drifts, or persistent improvements in split criteria. Ensembles present a complementary risk: even when replacement mechanisms of individual trees stabilize their long-run size, maintaining multiple learners and drift detectors can impose a large initial footprint~\cite{ADWIN_bifet2007learning,ARF_gomes2017adaptive}. Last, and maybe most severe, short experiments on only a few thousand data items may conflate online update capability with sustained deployability, overlooking either immediate infeasibility or growth over the deployment horizon.
% \begin{figure}
%     \centering
%     \includegraphics[width=1.0\linewidth]{Figures/best_base_learners_overtime_susy_8196 KB}
%     \caption{Avg accuracy and model size for simple base learners when memory is unlimited on susy dataset}
%     \label{fig:HT_growth}
% \end{figure}
While memory-aware stream learning is not absent from the literature (see e.g. ~\cite{da2018strict,buschjager2022shrub,Koebschall/etal/2026}), resource usage is still often a reported statistic or an implementation safeguard rather than a first-class experimental constraint. The software ecosystem reflects the same separation. MOA~\cite{bifet2018machine}, CapyMOA~\cite{gomes2025capymoa}, and River~\cite{montiel2021river} provide convenient research interfaces for prediction and incremental updates, but do not define a common, portable interface through which every learner can report its size, accept a changing budget, and actively enforce that budget. This matters for embedded deployment as much as for deployment on servers and the development of novel methods.

In this paper, we study the current state-of-the-art of stream-learning on long-horizon streams with explicit memory constraints. Our benchmark protocol is motivated by MCU-class deployment constraints, where memory available to a learner is often limited to kilobytes or a few megabytes. We evaluate representative online classifiers on seven real-world and six synthetic streams ranging from approximately 14 thousand to 11 million examples. The protocol combines prequential predictive performance with model size, update time, prediction time, and compliance with fixed memory budgets. Our goal is not to introduce a new learner, but to characterize the empirical baseline under a setting that stream learning was originally designed to address: sustained online learning under finite memory. Importantly, we do not claim to benchmark execution on a particular MCU, since such a comparison is currently difficult to perform systematically. We therefore tackle the prerequisite research question of whether current algorithms and their current reference implementations exhibit memory behavior compatible with MCU-class devices. Most importantly, the lack of MCU-capable streaming frameworks is not merely a matter of having the right software framework, but, as this work identifies, a matter of designing MCU-ready learning algorithms that can handle hard constraints. As a way forward, we present a potential API design that treats resource consumption as a primary design target and can serve as a basis for the next generation of stream learning algorithms.

%widely used stream-learning frameworks do not provide MCU deployment backends, a common representation-independent definition of model size, or an interface through which learners can enforce a memory budget. 
% Our contributions are as follows:
% \begin{itemize}
%     \item We formulate long-horizon memory-constrained stream learning as an empirical evaluation setting motivated by near-sensor, MCU-class deployments, where online learners are assessed under fixed model-size budgets over millions of updates.

%     \item We define a benchmark protocol that combines prequential predictive performance with peak model size, time to budget exhaustion, processing latency, and budget compliance.

%     \item We conduct a baseline study on seven real-world and six synthetic streams, spanning approximately 14K to 11M instances, covering representative incremental learners, ensembles, and memory-aware baselines.

%     \item We identify immediate and delayed memory failures, quantify long-term growth and latency degradation, and derive requirements for embedded-ready algorithms, evaluations, and software interfaces.
% \end{itemize}

Our contributions are as follows: we (i) identify a missing abstraction between stream-learning algorithms and resource-constrained deployment; %: current frameworks provide neither portable resource accounting nor enforceable budget contracts;
(ii) define a framework-independent evaluation protocol that approximates deployment readiness using initial footprint, peak model size, budget exhaustion, failure-aware accuracy, and latency development; (iii) apply this protocol to seven representative learners and expose two distinct resource-failure modes; and (iv) derive a concrete resource-aware learner interface intended to make future evaluation and MCU deployment systematic.

%Our contributions are as follows: we (i) formulate long-horizon memory-constrained stream learning as an evaluation setting for MCU-class deployments, (ii) define a benchmark combining accuracy, peak model size, budget exhaustion time, latency, and budget compliance, (iii) evaluate seven representative learners on 13 real and synthetic streams ranging from 14K to 11M instances, and (iv) identify memory-failure modes and derive requirements for embedded-ready algorithms, evaluations, and software interfaces.

%Throughout the paper, \emph{model size} means the learner-reported memory used by its predictive data structures at a measurement checkpoint. It excludes the language runtime, benchmark process, input buffers, and application memory. We use this narrower quantity because it can be compared across learners and is the resource that their learning mechanisms directly control. Total deployment RAM remains an important target for future on-device evaluation.

% but the experiments are performed on standard benchmarking hardware to enable large-scale and reproducible evaluation across many learners, streams, and seeds. 

\section{Related Work} %Stream Learning for Near-Sensor Platforms
%\todo[inline]{H: this section has a motivation around MCU, but the work discussed only covers streaming algorithm papers, none of which are about actually deploying online learners on microcontrollers. {we could include some MCU examples even if not stream? Maybe \url{https://arxiv.org/pdf/2103.08295}}. I've added it at the end of the section}

\begin{table}
\centering
\scriptsize
\setlength{\tabcolsep}{4pt}
\renewcommand{\arraystretch}{1.08}
\caption{Representative stream-learning methods.% All listed works evaluate predictive performance.
We distinguish whether concept drift, memory constraints, and latency are targeted by design or reported in the empirical evaluation.\cmark{} = yes; \pmark{} = partial/secondary; \xmark{} = no}
\label{tab:stream-resource-survey}

\begin{tabular*}{\linewidth}{
@{\extracolsep{\fill}}
l
ccc
ccc
@{}
}
\toprule
\multirow{2}{*}{Method}
& \multicolumn{3}{c}{Design}
& \multicolumn{3}{c}{Report} \\
\cmidrule(lr){2-4}
\cmidrule(lr){5-7}
& Drift
& Mem.
& Lat.
& Drift
& Mem.
& Lat. \\
\midrule

\shortstack[l]{HT/VFDT\\[-1pt]{\tiny\cite{domingos2000mining}}}
& \xmark
& \pmark
& \cmark
& \xmark
& \pmark
& \pmark \\

\shortstack[l]{CVFDT\\[-1pt]{\tiny\cite{hulten2001mining}}}
& \cmark
& \pmark
& \cmark
& \cmark
& \pmark
& \pmark \\

\shortstack[l]{HAT\\[-1pt]{\tiny\cite{bifet2009adaptive}}}
& \cmark
& \pmark
& \pmark
& \cmark
& \cmark
& \cmark \\

\shortstack[l]{ARF\\[-1pt]{\tiny\cite{ARF_gomes2017adaptive}}}
& \cmark
& \xmark
& \pmark
& \cmark
& \cmark
& \cmark \\

\shortstack[l]{EFDT\\[-1pt]{\tiny\cite{manapragada2018extremely}}}
& \xmark
& \xmark
& \xmark
& \cmark
& \xmark
& \xmark \\

\shortstack[l]{SVFDT\\[-1pt]{\tiny\cite{da2018strict}}}
& \pmark
& \cmark
& \cmark
& \cmark
& \cmark
& \cmark \\

\shortstack[l]{SRP\\[-1pt]{\tiny\cite{SRP_Gomes2019streaming}}}
& \cmark
& \pmark
& \xmark
& \cmark
& \cmark
& \cmark \\

\shortstack[l]{CS-ARF\\[-1pt]{\tiny\cite{bahri2020cs}}}
& \pmark
& \cmark
& \pmark
& \cmark
& \cmark
& \cmark \\

\shortstack[l]{GAHT\\[-1pt]{\tiny\cite{garcia-martin2021green}}}
& \pmark
& \cmark
& \cmark
& \cmark
& \cmark
& \cmark \\

\shortstack[l]{Shrubs\\[-1pt]{\tiny\cite{buschjager2022shrub}}}
& \pmark
& \cmark
& \cmark
& \cmark
& \cmark
& \cmark \\

\shortstack[l]{EFHAT\\[-1pt]{\tiny\cite{manapragada2022extremely}}}
& \cmark
& \xmark
& \xmark
& \cmark
& \xmark
& \xmark \\

\shortstack[l]{SGBT\\[-1pt]{\tiny\cite{Gunasekara2024}}}
& \cmark
& \xmark
& \pmark
& \cmark
& \cmark
& \xmark \\

\shortstack[l]{PLASTIC\\[-1pt]{\tiny\cite{heyden2024leveraging}}}
& \cmark
& \pmark
& \pmark
& \cmark
& \cmark
& \cmark \\

\bottomrule
\end{tabular*}
\end{table}

Stream learning has evolved around two central requirements: (i) maintaining high predictive performance under continuous data arrival and (ii) handling non-stationarity through concept-drift adaptation. However, as summarized in Table~\ref{tab:stream-resource-survey}, explicit consideration of resource constraints, particularly memory and latency, remains inconsistent across the literature. Foundational methods such as HT/VFDT established incremental tree induction~\cite{domingos2000mining}, while CVFDT, HAT, EFDT, and EFHAT introduced increasingly flexible mechanisms for revising model structure under concept drift~\cite{hulten2001mining,bifet2009adaptive,manapragada2018extremely,manapragada2022extremely}. Ensemble methods such as ARF, SRP, and SGBT further improve adaptation by combining multiple learners~\cite{ARF_gomes2017adaptive,SRP_Gomes2019streaming,Gunasekara2024}. Although these methods are computationally incremental and often report memory or runtime, they generally do not treat resource consumption as a primary design objective.

Several methods address this limitation more directly. SVFDT restricts tree growth, CS-ARF and GAHT incorporate resource-related objectives, and Shrubs explicitly bounds its ensemble through updating, pruning, and replacement~\cite{da2018strict,bahri2020cs,garcia-martin2021green,buschjager2022shrub}. PLASTIC similarly targets compact and efficient adaptation~\cite{heyden2024leveraging}. These works demonstrate that resource-aware stream learning is possible, but they remain exceptions within a literature dominated by predictive performance and drift adaptation. Moreover, resource awareness varies substantially: limiting growth or reducing average consumption does not necessarily guarantee that a learner will respect a hard memory budget under peak utilization or throughout an arbitrarily long stream.

TinyML and TinyOL approach the same constraint from a different direction. They demonstrate online neural adaptation on microcontrollers using restricted update schemes together with quantization, pruning, recomputation, paging, and hardware-specific implementations~\cite{Ren2021,Khouas2024,Zhu2024,Pavel2026,Velasquez2025} usually resulting in a specialized model architecture, training algorithm, and implementation for a given hardware, which are not generally applicable across devices and datasets.

Looking at existing stream learning frameworks, we perceive a lack of support for resource management. As Table~\ref{tab:frameworks} shows, MOA \cite{bifet2018machine}, CapyMOA \cite{gomes2025capymoa}, and River \cite{montiel2021river} currently (July 2026) support prediction and incremental updates as first-class abstractions, but not resource control: some learners expose budget parameters (e.g. \texttt{max\_memory}), yet no framework shares a common interface across all implemented methods for reporting model size. The definition itself also differs by framework. MOA measures model size by walking the full JVM object graph via a dedicated instrumentation agent, the \texttt{sizeofag} companion jar, including transient data structures. CapyMOA inherits this only partially: it combines methods implemented natively in Python with others that run on the JVM, so a single agent-based measurement cannot cover both, ruling out a uniform accounting strategy across the framework. River instead manually estimates model-state size with a fixed overhead factor, consistent with a library whose design centers on the API and usability rather than resource accounting. No framework offers a global memory budget enforced throughout execution, including management and evaluation overhead, and enforcement mechanisms (e.g., pruning) remain method-dependent and only partially supported. Last, budget failures are not handled at all. Although some methods do account for memory, this support is not consistent across frameworks. We see the lack of a common interface and evaluation protocol as a major bottleneck for developing and deploying stream learners on MCUs. %Table~\ref{tab:stream-resource-survey} maps the literature of available methods and shows that memory and latency are usually a secondary concern rather than a primary design target.

\begin{table}[]
\caption{Stream-learning frameworks and their feature set .\cmark{} = yes; \pmark{} = support depends on the selected method; \xmark{} = no}
\label{tab:frameworks}
\begin{tabular}{@{}lrrrr@{}}
\toprule
Capability & MOA & CapyMOA & River \\ \midrule
Accept budget & \pmark & \pmark & \pmark  \\
Report resource usage & \xmark & \xmark & \xmark \\
Enforce budget & \pmark & \pmark & \pmark \\
Expose budget failures & \xmark & \xmark & \xmark \\
MCU backend & \xmark & \xmark & \xmark  \\ \bottomrule
\end{tabular}
\end{table}

\section{Experimental Evaluation}

Our objective is to characterize the best behavior that each learner can attain under a memory budget, and then determine whether that behavior remains deployable over a long stream. We therefore distinguish three quantities that are often conflated: predictive quality, the learner's initial memory footprint, and subsequent model growth.
Our benchmark selects HT, HAT, EFDT, PLASTIC, ARF, SRP, and Shrubs because mature implementations are available, and they cover the principal families of incremental, adaptive, ensemble-based, and explicitly compact stream learners. Each learner is evaluated under various hyperparameter configurations and memory budgets to simulate different MCUs.

\subsection{Evaluation Protocol}

We use prequential evaluation: every example is first predicted and then used for one update. Hyperparameter optimization is intentionally optimistic. For each method, dataset, and budget, we select the highest-accuracy run among the evaluated configurations that satisfies the budget in hindsight. Consequently, this per-dataset oracle selection is not an estimate of tuning generalization, but the best observed outcome under favorable configuration choices. A configuration is budget-compliant only when the maximum recorded model size never exceeds the configured budget over the entire stream.
%Consequently, a resource limitation that remains under this protocol cannot be attributed merely to an unfortunate configuration choice.
We run HT, HAT, EFDT, PLASTIC, ARF, and SRP through CapyMOA's Python interface to their Java MOA implementations~\cite{gomes2025capymoa}. Shrubs uses a C++ implementation through a thin Python binding.

\subsection{Datasets}
\begin{table}
\centering
\scriptsize
\setlength{\tabcolsep}{2.5pt}
\renewcommand{\arraystretch}{1.02}
\caption{Benchmark streams grouped by type (synthetic in the top row, real-world in the bottom row). $N$, $d$, and $C$ denote instances, features, and classes. A dash denotes unknown natural drift.}
\label{tab:datasets}

\begin{tabular*}{\columnwidth}{@{\extracolsep{\fill}}l r r r l@{}}
\toprule
Dataset & $N$ & $d$ & $C$ & Drift \\
\midrule
%\multicolumn{5}{l}{\textit{Synthetic}} \\
AGR$_a$ & 10M & 9 & 2 & Abrupt (39 drifts) \\
AGR$_g$ & 10M & 9 & 2 & Gradual (39 drifts) \\
LED$_a$ & 10M & 24 & 10 & Abrupt (39 drifts) \\
LED$_g$ & 10M & 24 & 10 & Gradual (39 drifts) \\
RBF$_f$ & 10M & 10 & 5 & Fast incremental (3 drifts) \\
RBF$_m$ & 10M & 10 & 5 & Moderate incremental (3 drifts)  \\
\midrule
%\multicolumn{5}{l}{\textit{Real-world}} \\
Electricity & $\approx$45K & 8 & 2 & --- \\
Airlines & $\approx$539K & 7 & 2 & --- \\
HIGGS & 11M & 28 & 2 & --- \\
SUSY & 5M & 18 & 2 & --- \\
Weather & $\approx$18K & 8 & 2 & --- \\
Covtype & $\approx$581K & 54 & 7 & --- \\
Gas Sensor & $\approx$14K & 128 & 6 & Abrupt (6 drifts) \\
\bottomrule
\end{tabular*}
\end{table}
We evaluate the considered stream learners on a diverse set of real-world and synthetic data streams. The benchmark includes both binary and multi-class classification problems, as well as different types of concept drift (abrupt, gradual, and incremental).  Table~\ref{tab:datasets} summarizes the key characteristics of the datasets, including whether they exhibit drift, their origin, number of instances, features, and classes.
\begin{table}%[t]
\centering
\caption{Hyperparameter ranges for each method. Any other parameters are left to their respective defaults.}
\label{tab:hyperparameters}
\scriptsize
\setlength{\tabcolsep}{3pt}
% \begin{tabularx}{\columnwidth}{@{}l X@{}}
% \toprule
% Range & Parameters and methods \\
% \midrule
% $G=\{50,100,200\}$ & \texttt{grace\_period}: HT, HAT, EFDT, PLASTIC \\
% $C=\{10^{-5},10^{-3},10^{-1}\}$ & \texttt{confidence}: HT, HAT, EFDT; \texttt{split\_confidence}: PLASTIC \\
% $E=\{15,30,60,100\}$ & \texttt{ensemble\_size}: ARF, SRP \\
% $F=\{0.3,0.6,1.0\}$ & \texttt{max\_features}: ARF, SRP \\
% $B=\{32,64,128,256\}$ & \texttt{batch\_size}: Shrubs \\
% $L=\{15,30,60,100\}$ & \texttt{l\_ensemble\_reg}: Shrubs \\
% $\eta=\{0.1,0.5\}$ & \texttt{step\_size}: Shrubs \\
% Fixed & Shrubs: depth 20, MSE, hard-L0 regularization, normalized weights \\
% \bottomrule
% \end{tabularx}
\begin{tabular}{@{}lll@{}}
\toprule
Parameter & Ranges & Methods \\ \midrule
\texttt{grace\_period} & $\{50,100,200\}$ & HT, HAT, EFDT, PLASTIC \\
\texttt{ensemble\_size} & $\{15,30,60,100\}$ & ARF, SRP, Shrubs \\
\texttt{max\_features} & $\{0.3,0.6,1.0\}$ & ARF, SRP \\
\texttt{batch\_size} & $\{32,64,128,256\}$ & Shrubs \\
\texttt{step\_size} & $\{0.1,0.5\}$ & Shrubs \\
\texttt{max\_memory} & \begin{tabular}[c]{@{}l@{}}$\{128, 256, 512, 1024$, \\ $2048, 4096, 8192\}\text{KiB}$ \end{tabular} & All \\ \bottomrule
\end{tabular}
\end{table}
The six synthetic streams contain 10 million examples each. AGR and LED use matched abrupt and gradual transition schedules, while RBF provides fast and moderate incremental drift. For AGR and LED, we introduce 39 drifts (i.e. 40 concepts) evenly spaced every 250,000 items to account for the length of these datasets. The seven real-world streams range from roughly 14,000 to 11 million examples and cover binary and multi-class tasks with natural, unlabelled non-stationarity. %Together they vary in length, feature count, class count, and drift mechanism without assigning uncertain drift timestamps to the real-world data.

\subsection{Scope and Measurement Boundary}
No common embedded deployment stack currently exists for the considered methods as discussed previously. We therefore cannot compare compiled MCU memory footprints without independently reimplementing every learner. Since our goal is to assess MCU-readiness to guide such future implementations, we utilize CapyMOA and Shrubs' own C++ implementation. It is noteworthy that both implementations use two different memory accounting techniques. For the MOA learners, model size is obtained from \texttt{measureByteSize()}, which iterates the Java Object Hierarchy to determine its size. For Shrubs, it is obtained from the implementation's \texttt{num\_bytes()} method, which counts the size of the ensemble and its data structures manually. Absolute byte-level comparisons across both runtimes should therefore be interpreted cautiously. 
However, we are not interested in the specific memory consumption of a learner, as this is implementation- and runtime-dependent. We are interested in the algorithm-owned model state reported by the respective reference implementations. These measurements answer whether the \emph{learning algorithm} itself remains compatible with a given memory budget and show its growth factors and failure modes rather than exact equivalence between Java and C++ allocations. Put differently, our goal is not to compare ``good'' and ``bad'' implementations of the same learner, but to assess if a learning algorithm can potentially be deployed to an MCU with limited resources.

%On a more practical note, for the MOA learners, model size is obtained from \texttt{measureByteSize()} which iterates the Java Object Hierarchy to determine its size. For Shrubs, it is obtained from the implementation's \texttt{num\_bytes()} method, which counts the size of the ensemble and its data structures manually. Both estimate the learner's predictive structures rather than total Python, JVM, or C++ process memory. %This makes the measurements more comparable across implementations, but they remain implementation-reported estimates rather than allocator-level measurements.

\subsection{Hyperparameter and Run Coverage}

In our evaluation, we target modern 32-bit and AI-capable MCUs with model-size thresholds from 128\,KiB to 8,192\,KiB. Whenever a method exceeds the given budget during a stream, we stop training it but keep evaluating its performance until the stream ends. This simulates the arguably most direct and graceful way to address resource-usage during a long-running stream in which a learner grows over time. However, methods can still fail if their initial configuration cannot be deployed under a given memory budget at all. In addition, we execute one unconstrained run that does not impose any limits on the method. For each threshold, we randomly sampled up to 20 configurations per method following the random search optimization by \cite{bergstra2012random} and repeat each configuration three times with different random seeds. Table~\ref{tab:hyperparameters} give an overview of the hyperparameter ranges. For each run, we set a total limit of 12h, stopping and discarding results whenever the run did not finish within that time. All learners were configured to utilize a single core because single-core CPUs are dominant in MCUs. All experiments were conducted on multiple compute nodes that allocated one AMD EPYC 7742 CPU with 3 GB RAM per configuration. Out of a total of 6,463 experiments, 6,109 executed successfully, and 354 failed because of memory limit, execution, or infrastructure factors. Figure~\ref{fig:experiment-inventory} shows how these runs are distributed across methods. It is noteworthy that SRP and ARF are the only methods with failed runs. These occurred partially because these methods would require more memory in the beginning than the budget allowed and sometimes consume more than the configured 3 GB of RAM or because their time limit was exhausted.

\begin{center}
\begin{minipage}{\linewidth}
    \centering
    \includegraphics[width=0.9\linewidth]{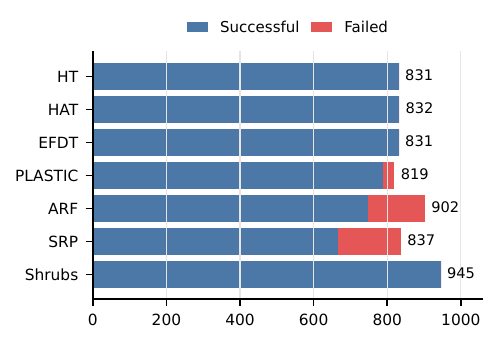}
    \captionof{figure}{Distribution of successful and failed experiments.}
    \label{fig:experiment-inventory}
\end{minipage}
\end{center}

\section{RQ1: Accuracy Under a Fixed Budget}

\emph{Question.} Which method attains the best predictive performance under a fixed memory budget? \\
\emph{Hypothesis.} Most methods in Table~\ref{tab:stream-resource-survey} focus on concept drift rather than hard memory constraints. Moreover, HT, which is often used as an ensemble base learner, tends to grow as more data arrive. We therefore hypothesize that compact or explicitly bounded learners rank best at small budgets, whereas adaptive ensembles recover their accuracy advantage as the budget increases.\\
\emph{Analysis.} To answer this question, we select the most accurate eligible configuration for each dataset, method, and target budget. If a completed sweep contains no eligible configuration, its accuracy is set to zero for ranking, representing a learner that cannot be deployed under that budget. Figure~\ref{fig:rq1-ranks} is a stack of seven critical-difference (CD) diagrams \cite{demsar} where each row represents one memory budget. Within a row, each colored marker is a method's average rank across the 13 datasets, where rank 1 (more to the left) is best. Black comparison bars join methods that are not statistically distinguishable at $\alpha=.05$ after Holm correction. A detailed accuracy comparison can be found in the appendix. \\
%The display is therefore two-dimensional only in layout: the vertical direction enumerates budgets and is not a second numerical axis.
\emph{Finding.} Our original hypothesis is supported and the findings are largely as expected. At 128\,KiB, neither ARF nor SRP produces a peak-compliant configuration, whereas PLASTIC and Shrubs are compliant on all 13 datasets and attain the leading ranks. With more memory, ARF and SRP move toward the best ranks. For smaller budgets below 1 MB, Shrubs, HT, and EFDT are generally the best methods. For larger budgets, ARF and SRP dominate, followed by Shrubs in third place.

\begin{center}
\begin{minipage}{\linewidth}
    \centering
    \includegraphics[width=0.9\linewidth]{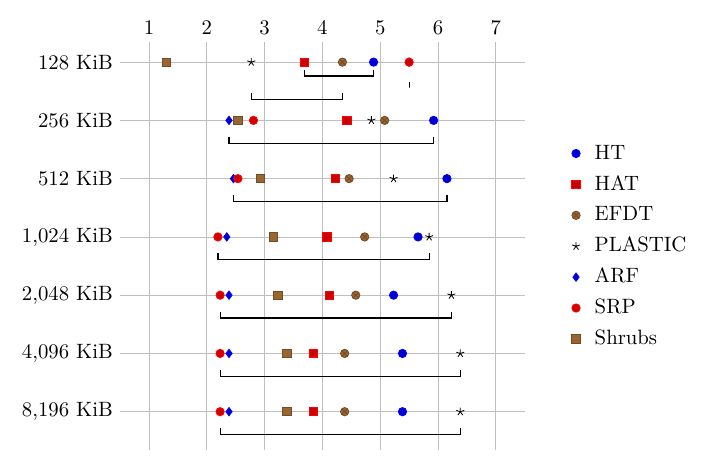}
    \captionof{figure}{Ranks by model-size budget. Each row is an independent CD diagram where lower average ranks are better.}
    \label{fig:rq1-ranks}
\end{minipage}
\end{center}

\section{RQ2: Long-Horizon Model Growth}

\begin{figure}%[t]
    \centering
    \includegraphics[width=.9\columnwidth]{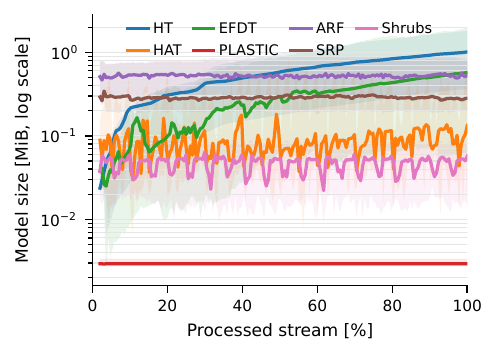}
    \\[-0.2em]
    \smallskip
    {\small\textbf{(a)} Model-size trajectories over normalized stream progress.\par}
    \medskip
    \includegraphics[width=.9\columnwidth]{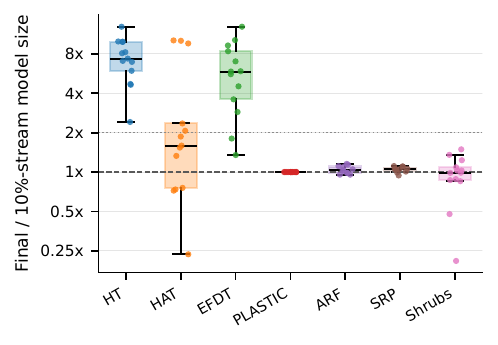}
    \\[-0.2em]
    \smallskip
    {\small\textbf{(b)} Dataset-level growth between 10\% and the end of the stream.\par}
    \caption{Growth of the highest-accuracy unconstrained configuration for each method and dataset. Panel (a) shows the progression of model size over the stream. Panel (b) shows the growth ratio per method.}
    \label{fig:rq2-growth}
\end{figure}

\emph{Question.} Which learners continue to grow over a long stream, and how quickly? \\ 
\emph{Hypothesis.} HT, HAT, and EFDT extend their tree topology and rarely remove existing structures, so we expect them to grow over time. ARF and SRP use these trees as base learners, although they manage them more dynamically. We hypothesize that model size for all five methods grows materially with stream length and that the tree ensembles grow at least as much as individual trees. Explicitly bounded methods should instead stabilize. \\
\emph{Analysis.} For this analysis, we only consider the best average (over seeds) unconstrained configuration for each method. This produces one trajectory for each dataset and method. We express progress as a percentage of the dataset length so that streams of different lengths can be compared. At every percentage point, Figure~\ref{fig:rq2-growth}(a) shows the median model size across the dataset trajectories. The shaded area contains the middle 50\% of datasets. Note the logarithmic axis. Figure~\ref{fig:rq2-growth}(b) summarizes growth within each dataset: We first measure the model size after 10\% of each stream and use this as a normalization for the final model size. Starting at 10\% avoids treating initial allocation and one-time initialization as sustained growth. A value of $1\times$ means no net growth and $2\times$ means that the model doubled in size. The box plots show the distributions over all datasets and random seeds. \\
\emph{Finding.} Our hypothesis is only partially supported. HT, EFDT, and HAT behave as expected: HT grows by a median factor of 7.37 between 10\% and 100\% of a stream, and at least doubles on all 13 datasets. EFDT grows by $5.87\times$ and doubles on 11 of 13 datasets. HAT is heterogeneous. Its median growth is $1.59\times$, with five datasets above $2\times$. Contrary to our hypothesis, ARF and SRP start much larger but have median growth factors of only 1.04 and 1.05. PLASTIC and Shrubs remain constant as expected. 
The small relative growth of ARF and SRP, albeit unexpected, is plausible for two reasons. First, both methods allocate a fixed number of ensemble members, so their large initial footprint is already present early in the stream. Second, drift adaptation can replace or reset existing trees instead of adding new members indefinitely. Removing an old tree also discards its accumulated structure and can offset growth elsewhere in the ensemble. Memory risk therefore takes two forms. ARF and SRP can be infeasible initially, while HT and EFDT may become infeasible only after sustained use.

\section{RQ3: Latency Degradation}

\begin{figure}%[t]
    \centering
    \includegraphics[width=0.9\linewidth]{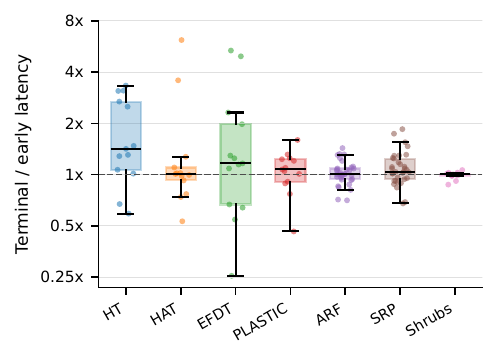}
    \caption{Distribution of slowdowns.}
    \label{fig:rq4-slowdown}
\end{figure}

\emph{Question.} Do computational costs increase over a long stream? \\
\emph{Hypothesis.} RQ2 shows that HT and EFDT grow, while the aggregate size of ARF and SRP remains comparatively stable. We therefore expect latency degradation for HT and EFDT and little relative change for methods whose size remains stable. \\
\emph{Analysis.} We analyze the same highest-accuracy unconstrained configurations used in RQ2, but measure test-then-train latency for each example. Similar to RQ2, we ignore the first part of the stream to obtain a reference measurement because the first observations may include one-time initialization and unusually small model structures, so they are not a stable reference. For each dataset, learner, and seed, we instead define early-stream latency as the median between 2\% and 10\% of the stream and terminal latency as the median over the final 10\%. Dividing terminal by early-stream latency gives the slowdown ratio shown in Figure~\ref{fig:rq4-slowdown}. The dashed $1\times$ line denotes unchanged latency. Values above $1\times$ indicate slowdown, while values below $1\times$ indicate faster processing. 
\emph{Finding.} The hypothesis is partially supported. HT has a median slowdown of $1.42\times$, while EFDT reaches $1.16\times$. Their broad distributions show that model growth does not translate into the same amount of additional work on every dataset. Hence, model size is only an indirect measure of latency. The remaining methods have a median slowdown close to $1\times$, with some notable outliers for HAT and PLASTIC. We hypothesize that this is due to the restructuring of the trees, which can be triggered in both methods. ARF and SRP also have some outliers, but much less severe, while Shrubs has basically a constant latency. 

% HAT has a median ratio close to $1\times$ with some notable outliers. A plausible explanation is its adaptive-tree mechanism. Alternate subtrees and branches can be discarded or replaced after detected change. A newly selected subtree can be smaller or shallower than the structure it replaces, which reduces traversal and update work. 
% % This interpretation agrees with the heterogeneous HAT growth in RQ2 and the repeated contraction and regrowth visible in RQ5. It does not establish that every latency reduction was caused by a drift response because the benchmark does not directly align individual replacement events with latency measurements.
% ARF and SRP also remain close to $1\times$. Both methods maintain a fixed number of ensemble members and primarily replace existing trees instead of continually adding new ones. Their stable latency is therefore consistent with the small aggregate growth factors observed in RQ2. RQ3 nevertheless shows that ARF can violate small memory budgets almost immediately. A large initial footprint can prevent deployment even when latency does not degrade further over time. PLASTIC and Shrubs are similarly stable, with Shrubs showing a median of $1.00\times$ and an IQR of 0.99--1.01. These results connect memory growth to latency degradation without treating memory as a direct substitute for computational cost.

\section{RQ4: Time to Budget Exhaustion}
\begin{figure*}
    \centering
    \includegraphics[width=0.8\textwidth]{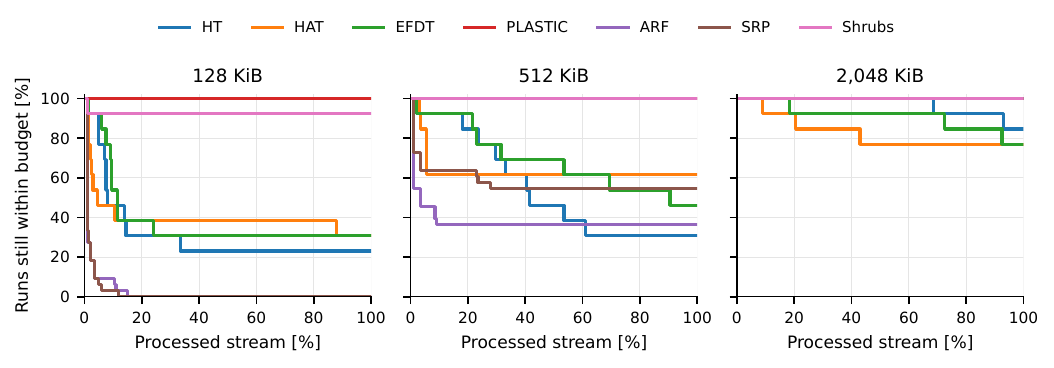}
    \caption{Percentage of accuracy-optimal configurations that remain within the specified budget over normalized stream progress.}
    \label{fig:rq3-survival}
\end{figure*}

\emph{Question.} How long does an accuracy-optimal learner remain within a fixed memory budget? \\ 
\emph{Hypothesis.} Motivated by RQ2, we hypothesize two distinct exhaustion modes: large ensembles can be infeasible almost immediately because of their initial footprint, whereas initially small trees cross a budget later as their structure grows. Increasing the budget should delay or eliminate both modes. \\
\emph{Analysis.} We reuse the accuracy-optimal unconstrained configurations from RQ2 and apply each candidate budget to their recorded progressions over the stream after the fact. In each panel of Figure~\ref{fig:rq3-survival}, the horizontal axis is normalized stream progress, and the vertical axis is the percentage of accuracy-optimal configurations on each dataset that have not yet exceeded the stated budget. A curve that drops close to zero near the origin indicates initial infeasibility, while a gradual decline indicates exhaustion during operation. We focus on three budgets: 128\,KiB, 512\,KiB, and 2\,MiB as representative regimes. The appendix contains plots for all seven budgets. \\
\emph{Finding.} The hypothesis is supported. At 128\,KiB, the accuracy-optimal ARF and SRP configurations are almost uniformly infeasible from the beginning, quickly followed by HT, HAT, and EFDT. PLASTIC and Shrubs remain feasible as expected. At 512\,KiB, exhaustion occurs later in the stream but remains an issue on all methods except PLASTIC and Shrubs. At 2\,MiB, most learners complete the stream without exhausting the budget. Here, only HAT, EFDT and ARF still struggle, while most learners finish within the budget.

\section{RQ5: Model Size and Concept Drift}

\begin{figure}[h!]
    \centering
    \includegraphics[width=.8\columnwidth]{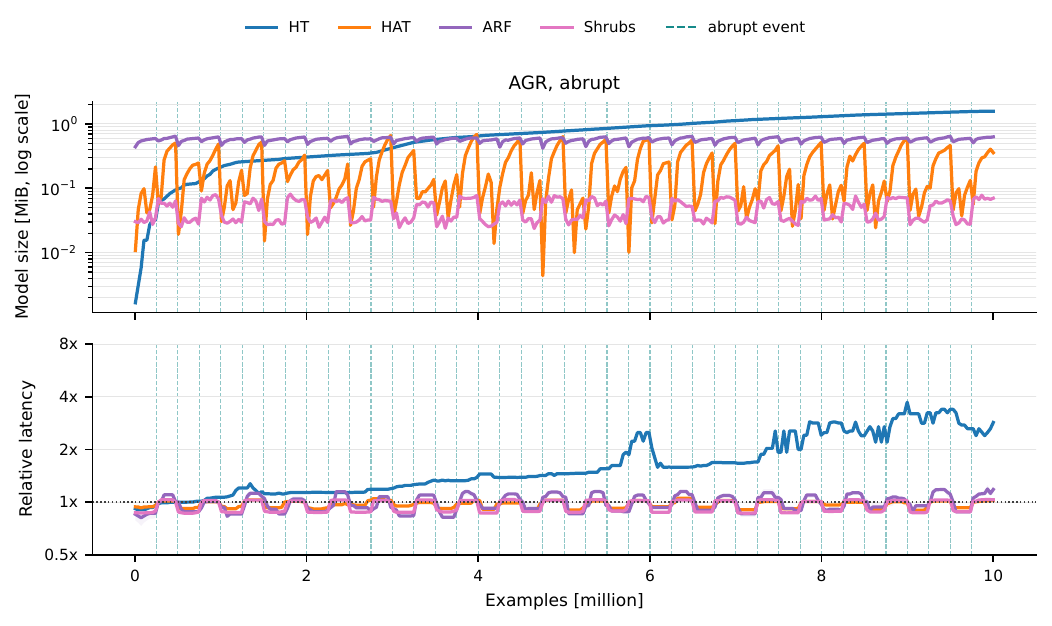}
    \\[-0.2em]
    \smallskip
    {\small\textbf{(a)} AGR, abrupt drift.\par}
    \medskip
    \includegraphics[width=.8\columnwidth]{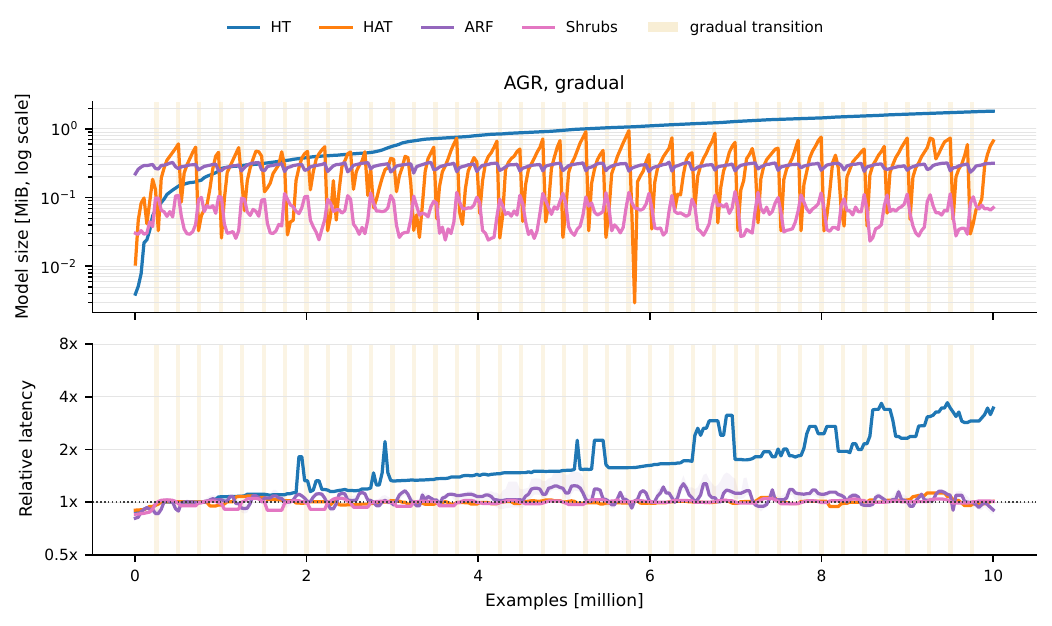}
    \\[-0.2em]
    \smallskip
    {\small\textbf{(b)} AGR, gradual drift\par}
    \caption{Model size and prediction-plus-update latency. Teal dashed lines mark abrupt events, and pale-gold
    intervals mark gradual transitions.}
    \label{fig:rq5-drift}
\end{figure}

\emph{Question.} How does model size behave around (recurring) drift events? \\
\emph{Hypothesis.} We hypothesize that adaptive learners show event-aligned contraction, replacement, or regrowth, whereas HT retains structure across transitions and Shrubs remains bounded. \\
\emph{Analysis.} We focus on the two synthetic datasets (AGR with abrupt and gradual drift) because their drift events are known. 
%This comparison tests whether event-aligned changes in model size also produce changes in processing cost. A shaded band extends from the 25th to the 75th percentile across seeds. 
For visibility, we use the highest-accuracy unconstrained configuration of each of four representative methods: HT, HAT, ARF, and Shrubs. They were selected to isolate different resource mechanisms. HT is a non-adaptive growing-tree baseline that retains learned structure. HAT adds branch-level adaptation and can replace accumulated structure. ARF tests whether a fixed-size adaptive ensemble of tree learners stabilizes size and latency through member replacement. Shrubs provides an explicitly memory-bounded ensemble baseline. Figure~\ref{fig:rq5-drift}(a) plots model size over the stream on a logarithmic axis. Figure~\ref{fig:rq5-drift}(b) plots test-then-train latency relative to the early median stream (obtained from the first 2\% - 10 \% of the stream).\\
%We inspect the accuracy-optimal unconstrained configurations on four 10-million-example AGR and LED streams with 39 transitions spaced 250,000 examples apart. 
%EFDT largely overlaps the single-tree growth mechanism represented by HT, SRP overlaps the adaptive-ensemble mechanism represented by ARF, and PLASTIC has an almost constant size similar to the bounded case. Size and latency results for all seven methods are reported in the appendix. Teal dashed lines identify abrupt events, while pale-gold regions show the 50,000-example windows over which gradual transitions occur.
\emph{Finding.} The model size progression supports the hypothesis. HT grows throughout all both streams without an event-driven reset. HAT repeatedly contracts and regrows, often near transition markers. ARF fluctuates around a comparatively stable operating level, which is consistent with the replacement of constituent models, while Shrubs remains compact and bounded. %These progressions also provide a mechanistic explanation for the RQ2 result that an ensemble of growing base learners need not show sustained aggregate growth. 
Looking at latency, we see that it separates persistent growth from adaptive replacement more clearly. HT becomes progressively slower as its retained structure increases. By contrast, the repeated contractions of HAT and Shrubs and the bounded fluctuations of ARF do not create a sustained latency trend. Their curves remain close to the early-stream baseline. Model size is therefore informative about long-term computational pressure, but event-scale size variation does not translate one-to-one into latency. 
% Without matched no-drift controls and direct logs of replacement events, the alignment remains descriptive and cannot establish that drift caused each individual change.

% \begin{figure*}%[h]
%     \centering
%     \includegraphics[width=.9\textwidth]{Figures/rq5_drift_response.pdf}
%     \\[-0.2em]
%     \smallskip
%     {\small\textbf{(a)} Model-size.\par}
%     \medskip
%     \includegraphics[width=.9\textwidth]{Figures/rq5_drift_latency.pdf}
%     \\[-0.2em]
%     \smallskip
%     {\small\textbf{(b)} Latency relative to the early-stream baseline.\par}
%     \caption{Model size and prediction-plus-update latency. Teal dashed lines mark abrupt events, and pale-gold
%     intervals mark gradual transitions.}
%     \label{fig:rq5-drift}
% \end{figure*}

\subsection{Discussion and Future Directions}

\begin{algorithm}%[t]
\caption{Resource-aware stream-learning loop.}
\label{alg:stream-learning-loops}
\begin{algorithmic}[1]
\STATE $\textsc{set\_budget}(B)$
\WHILE{next item $(x,y)$ exists}
    \STATE $\hat{y} \gets \textsc{predict}(x)$
    \STATE $\ell \gets \textsc{eval}(\widehat y, y)$
    \STATE $\textsc{report\_performance}(\ell)$
    \STATE $\hat{b} \gets \textsc{resource\_usage}()$
    \STATE $\textsc{report\_resource\_usage}(\widehat b)$
    \IF{$\hat{b}>B$}
        \STATE $\textsc{report\_failure}(\widehat b)$
        \STATE $\textsc{enforce\_budget}(\widehat b)$
    \ELSE
        \STATE $\textsc{train\_on\_instance}(x,y)$
    \ENDIF
    %\STATE $\textsc{enforce\_budget}(\hat b, B)$ 
\ENDWHILE
\end{algorithmic}
\end{algorithm}

Taken together, the five research questions show that being incremental or drift-adaptive does not make a learner resource-bounded. RQ1 identifies a clear accuracy--memory trade-off. Compact methods remain viable under small budgets, while large adaptive ensembles become competitive only when more memory is available. RQ2 and RQ4 reveal two distinct failure modes. ARF and SRP can be excluded almost immediately by their initial footprint even though their later footprint is comparatively stable. HT and EFDT can fit initially and then exhaust the same budget as the retained structure accumulates. RQ3 shows that this growth often, but not invariably, increases processing latency. RQ5 links these aggregate trends to learner mechanisms: retained trees grow, adaptive trees contract and regrow, and fixed-size or bounded ensembles operate around a more stable size. Our results expose two gaps. The first is algorithmic: many stream learners either begin above MCU-class budgets or grow beyond them over time. The second is infrastructural: current frameworks provide no common mechanism for expressing, enforcing, or testing a resource budget. 

%The absence of MCUs in the stream-learning literature is therefore not only an evaluation choice, but reproducing such results currently requires method-specific ports and memory accounting.
MCU-hardware evaluation remains uncommon in the stream-learning literature, in part because it currently requires method-specific ports and manual memory accounting rather than a shared measurement path. Closing the infrastructural gap is necessary before MCU deployment can become a routine benchmark rather than an isolated engineering project in stream learning. At minimum, evaluations should report initial footprint, peak model size, time to budget exhaustion, failure-aware predictive performance, and latency over sufficiently long streams. This turns memory and latency from an implementation detail into an explicit property of the learning problem that we refer to as \emph{resource-aware prequential evaluation}. More strongly, we advocate that stream learning should move from ``learning under concept drift'' to  ``learning under concept drift \emph{with limited resources}''. We propose to extend the typical test-then-train core loop of today's stream learning frameworks to directly include resources, for example, through an appropriate API as depicted in Algorithm~\ref{alg:stream-learning-loops}. Then, future work can effectively tackle the next step in stream learning: not only how to detect and adapt to change but also how to sustain that adaptation within a fixed resource envelope.

\bibliography{aaai2027}
 
\section{APPENDIX}
\onecolumn

\begin{figure*}
    \centering
    \includegraphics[width=0.8\textwidth]{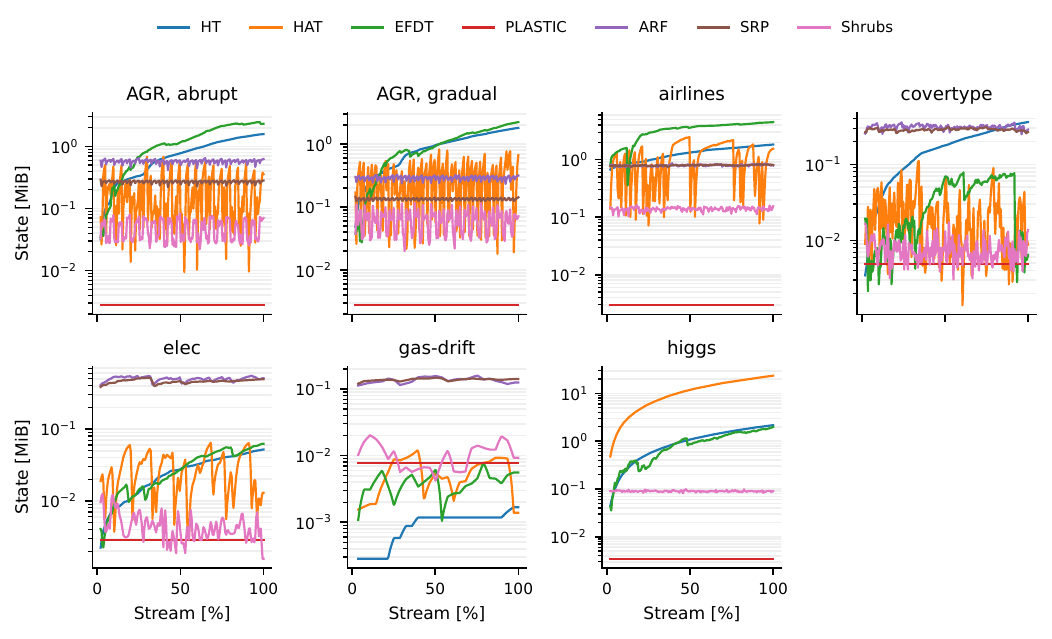}
    \caption{RQ2 model-size trajectories, datasets 1--7.}
    \label{fig:app-rq2-1}
\end{figure*}
\begin{figure*}
    \centering
    \includegraphics[width=0.8\textwidth]{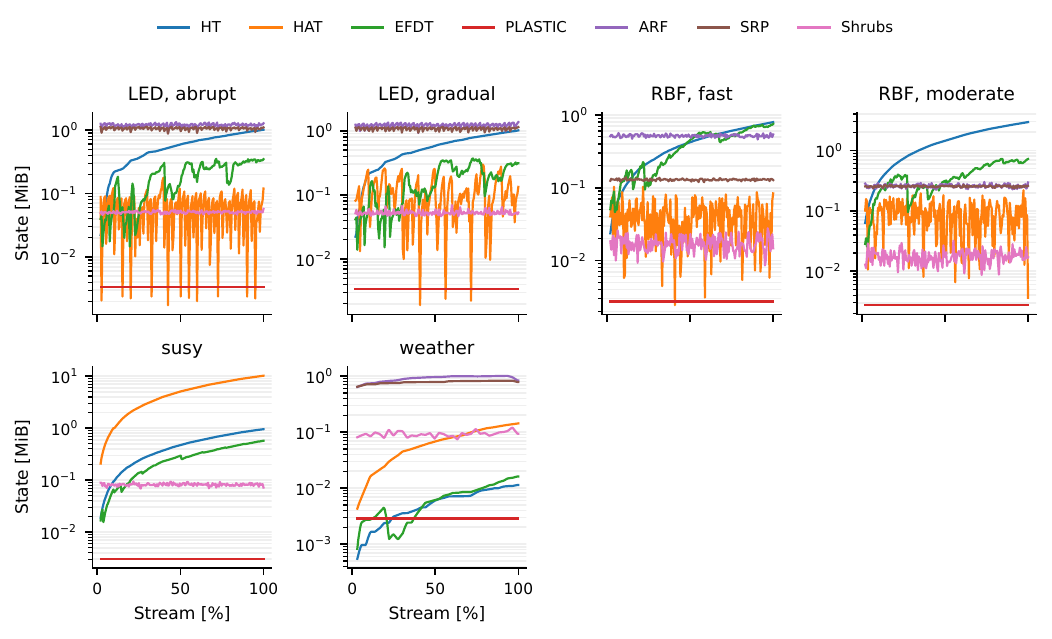}
    \caption{RQ2 model-size trajectories, remaining datasets.}
    \label{fig:app-rq2-2}
\end{figure*}

\begin{figure*}
    \centering
    \includegraphics[width=\textwidth]{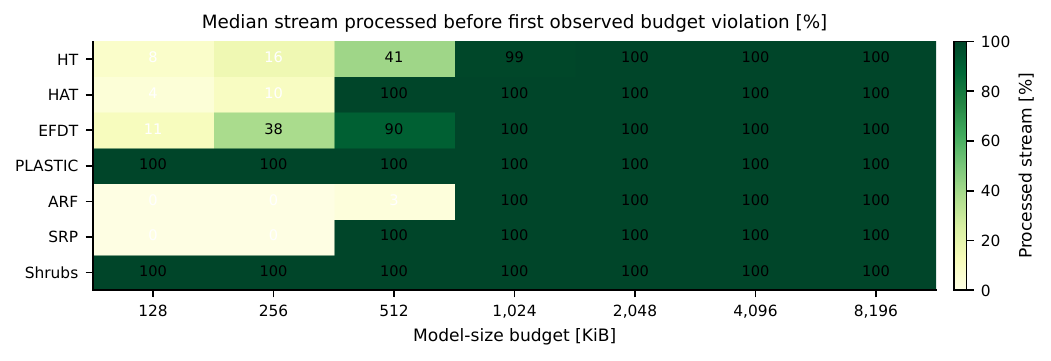}
    \caption{RQ3 median stream percentage before the first observed violation
    for every configured budget. A value of 100 means that the median trajectory
    completes, not that every trajectory is compliant.}
    \label{fig:app-rq3}
\end{figure*}

\begin{figure*}
    \centering
    \includegraphics[width=\textwidth]{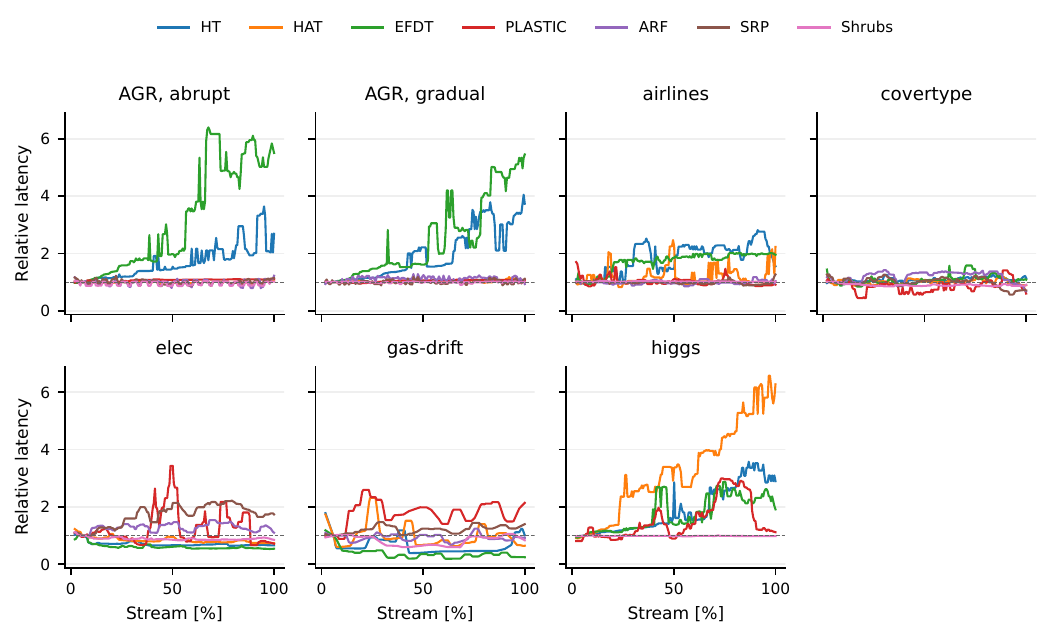}
    \caption{RQ4 relative-latency trajectories, datasets 1--7.}
    \label{fig:app-rq4-1}
\end{figure*}

\begin{figure*}
    \centering
    \includegraphics[width=\textwidth]{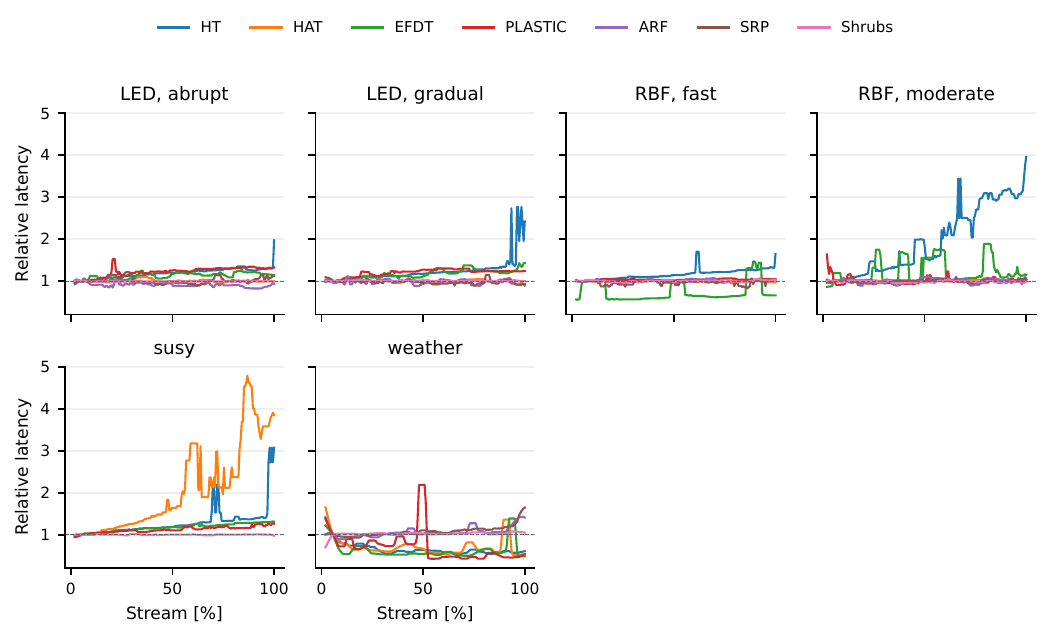}
    \caption{RQ4 relative-latency trajectories, remaining datasets.}
    \label{fig:app-rq4-2}
\end{figure*}

\begin{figure*}
    \centering
    \includegraphics[width=\textwidth]{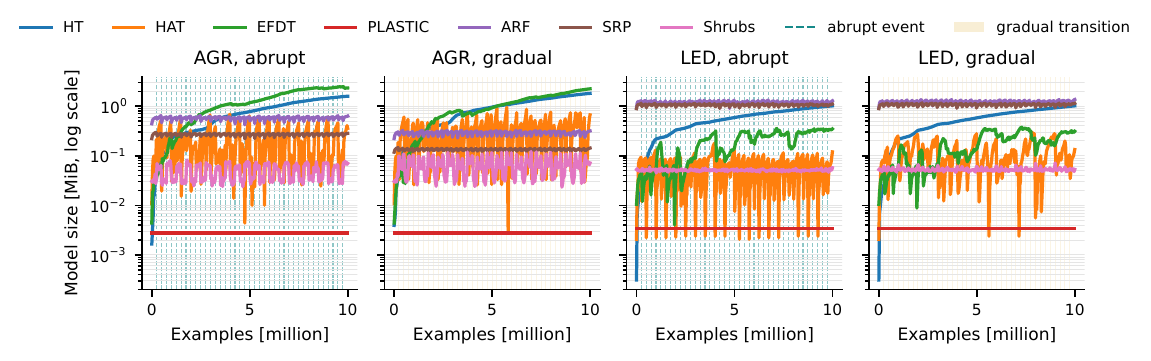}
    \caption{RQ5 model-size trajectories for all seven methods.}
    \label{fig:app-rq5}
\end{figure*}

\begin{figure*}
    \centering
    \includegraphics[width=\textwidth]{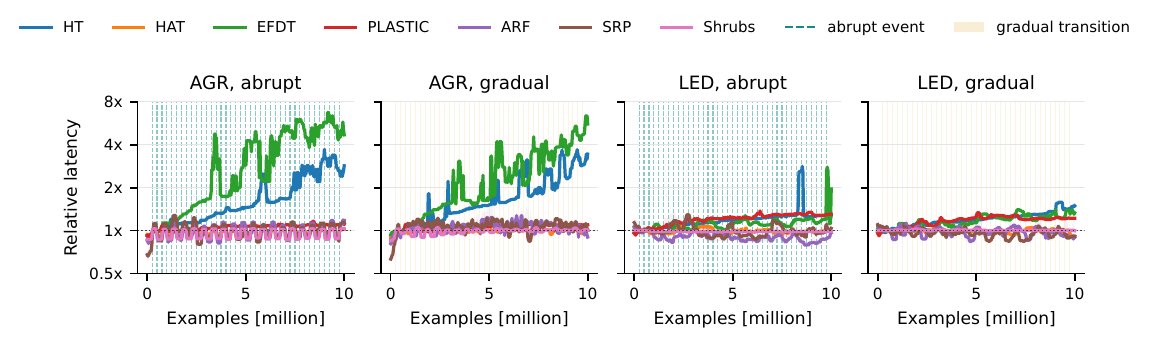}
    \caption{RQ5 relative-latency trajectories for all seven methods.}
    \label{fig:app-rq5-latency}
\end{figure*}


\end{document}